\documentclass[]{fairmeta_gr2}

\usepackage{colortbl}
\usepackage{mdframed}
\usepackage{fvextra}
\usepackage{algorithm}
\usepackage{algpseudocode}
\usepackage{amsthm}
\usepackage{amssymb}
\usepackage{amsmath}
\usepackage{mathtools}
\usepackage{xspace}
\usepackage{textcomp}
\usepackage{fancyvrb}

\microtypesetup{expansion=false}

\definecolor{doderblue}{RGB}{30,144,255}
\definecolor{lightblue}{RGB}{210, 220, 250}
\definecolor{tagblue}{RGB}{0, 0, 200}

\newcommand{\ReDiST}{\textsc{ReDiST}\xspace}

\title{Beyond the Answer: Student-Conditioned Trace Construction for Reasoning Distillation}

\author[1, *]{Shaoang Li}
\author[1, *]{Yanhang Shi}
\author[2]{Yufei Li}
\author[2]{Mingfu Liang}
\author[2]{Frank Shyu}
\author[2]{Sandeep Pandey}
\author[2]{Luke Simon}
\author[2, \dagger]{Xi Liu}
\author[1, \dagger]{Jian Li}

\affiliation[1]{Stony Brook University}
\affiliation[2]{Meta AI}

\contribution[*]{Joint First Author}
\contribution[\dagger]{Joint Correspondence Author}

\abstract{
A reasoning trace can accompany a correct answer without providing effective supervision for a particular student.  
We study how to construct traces that improve learning beyond answer-only supervision by considering both their intermediate content and their accessibility to the target learner. 
We introduce \ReDiST, a student-conditioned trace-construction and distillation framework for evidence-rich reasoning. Starting from the student's own attempts, \ReDiST revises answer-correct drafts against the supplied evidence and repairs unsuccessful drafts with additional guidance from the verified answer.
A frozen student-matched reader screens and ranks original and edited candidates using answer recovery and verified-answer probability, given the question, evidence, and candidate trace. We distill the selected targets while balancing trace and final-answer supervision with an auxiliary answer loss. 
Construction and selection occur offline, and inference uses only the distilled student. 
Experiments on question-answering benchmarks show that \ReDiST can improve both source-task and cross-benchmark performance over answer-only, original-trace, and the evaluated trace-selection and rewriting baselines. Across supervision recipes, performance rankings can differ between source and transfer benchmarks, highlighting the importance of evaluating trace teaching utility across distributions.
}

\date{October 16, 2026}

\begin{document}

\maketitle

\section{Introduction}
\label{sec:intro}

Reasoning-trace distillation trains a language model on intermediate derivations as well as final answers, making generated solutions a practical source of supervision \citep{hsieh2023distilling,naturalthoughts}. In evidence-rich tasks such as multi-hop question answering, a derivation can communicate supporting facts and intermediate relations that are useful on new questions. Yet the trace also becomes a sequence the student must learn to predict: missing reasoning steps, unnecessary elaboration, or unfamiliar expression can make it an ineffective teaching target even when its final answer is correct. The practical challenge is therefore \textit{to construct reasoning traces that improve the student's learning beyond answer-only supervision.}

\begin{figure}[h]
    \centering
     \vspace{-0.1in}
    \includegraphics[width=0.85\linewidth]{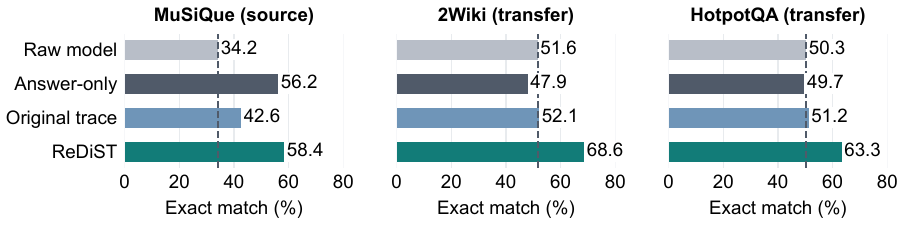}
    \caption{\textbf{Source-task accuracy and cross-benchmark transfer respond differently to supervision.} Qwen3-8B before task-specific training (Raw) and after training on MuSiQue with answer-only supervision, original student traces, or \ReDiST. MuSiQue measures source-task performance, while 2Wiki and HotpotQA measure cross-benchmark performance. Dashed lines mark each benchmark's raw-model EM. Values are EM point estimates from Table~\ref{tab:musique-transfer}. Comparisons between trained variants reflect complete training recipes.}
    \label{fig:distillation-motivation}
     \vspace{-0.1in}
\end{figure}

Figure~\ref{fig:distillation-motivation} makes this challenge concrete. With MuSiQue supervision, answer-only training raises Qwen3-8B's source-task exact match (EM) from 34.2\% to 56.2\%, while its EM on 2WikiMultiHopQA (2Wiki) falls from 51.6\% to 47.9\%. Training on the original student traces yields a different profile: 42.6\% source EM and 52.1\% transfer EM. The answer-only and original-trace recipes therefore rank differently on the two benchmarks: \textit{higher source-task accuracy does not necessarily imply better cross-benchmark performance.} These contrasting profiles motivate assessing trace supervision through both source-task and cross-benchmark performance.

Prior work improves reasoning supervision through self-training and correction \citep{star,rerest,cgd}, including metacognitive rewriting of student attempts \citep{mbt}. Other methods select trajectories for their informativeness and learnability \citep{rsr,lark}, preserve the student's style during refinement \citep{qualityutility}, or use student answer prediction to optimize explanations and select rationales \citep{rlt,coalition}. These studies establish the importance of both trace content and compatibility with the learner. Building on these insights, we focus on a specific construction problem in evidence-rich reasoning: \textit{deciding which evidence-guided revisions and repairs of the target student's own attempts should become its teaching targets.}
We couple evidence-guided editing with selection based on the frozen student's response, and examine what this construction pipeline adds beyond selecting unedited student drafts while accounting for the allocation of trace and answer supervision.

These observations motivate the following question:
\begin{quote}
    \emph{How can we turn the target student's own attempts into reasoning supervision that improves learning beyond final-answer labels?}
\end{quote}

We approach this question through two considerations. \textbf{Process content} concerns the supporting facts and intermediate relations communicated beyond the final-answer label, while \textbf{student accessibility} concerns the target student's ability to use and learn from that content. A conclusion-only trace can reveal the answer while omitting the reasoning that supports it, whereas a detailed derivation can supply that reasoning in a form the student struggles to learn. Improving a derivation's content therefore need not make it a better teaching target for a particular student.

\begin{figure}[t]
    \centering
    \includegraphics[width=\linewidth]{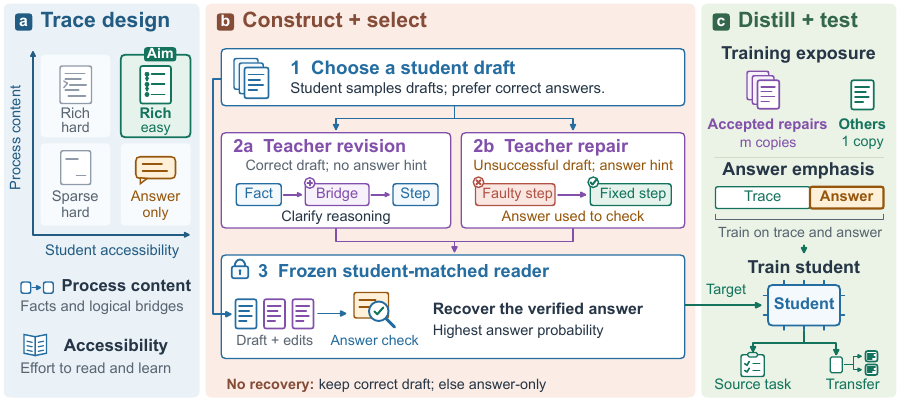}
    \vspace{-0.3in}
    \caption{\textbf{\ReDiST: student-conditioned trace construction and distillation.}
    (a) Process content and student accessibility guide the design.
    (b) A rewrite model revises answer-correct student drafts against the supplied evidence and repairs unsuccessful drafts with additional guidance from the verified answer. A frozen student-matched reader screens and ranks original and edited candidates using answer recovery and verified-answer probability, given the question, evidence, and candidate trace.
    (c) Repair exposure $m$ and auxiliary answer-loss weight $\alpha$ control how selected targets enter training. Teaching utility is evaluated through the trained student's source-task and cross-benchmark performance.}
    \label{fig:redist-overview}
     \vspace{-0.1in}
\end{figure}

To address these considerations, we introduce \ReDiST (\emph{Refinement and Distillation of Student Traces}), a student-conditioned trace-construction and distillation framework for evidence-rich reasoning (Figure~\ref{fig:redist-overview}). \textit{Its organizing principle is to let the target student shape both the starting drafts and the selection of teaching targets.} A rewrite model revises answer-correct drafts using the question, supplied evidence, and original attempt. For unsuccessful drafts, it additionally receives the verified answer as guidance for repairing the supporting derivation against the evidence. The two branches address complementary needs: \textit{improving intermediate supervision even when the student's final answer is correct, and recovering supervision for questions it initially fails to solve.}

A frozen student-matched reader then receives the question, evidence, and each original or edited candidate trace, and screens and ranks candidates using answer recovery and verified-answer score. This supplies a practical selection signal, while teaching utility is evaluated through the student's performance after distillation. During training, repair exposure controls how often accepted repairs are seen, while an auxiliary answer loss controls the emphasis on final-answer prediction alongside trace supervision. These choices separate which targets are constructed from how they contribute to optimization. Construction and selection occur offline, and inference uses only the distilled student.

Under MuSiQue supervision, \ReDiST improves source-task and 2Wiki performance over answer-only and original-trace supervision across four students from two model families (Table~\ref{tab:cross-setting-utility}). In the primary Qwen3-8B comparison, it achieves the highest reported point estimates across all four benchmarks, including 58.4\% MuSiQue EM and 68.6\% 2Wiki EM (Table~\ref{tab:musique-transfer}). A separate Qwen3-8B comparison extends these gains to a 2Wiki supervision and transfer to MuSiQue (Table~\ref{tab:2wiki-transfer}). Component comparisons show additional gains beyond selecting unedited student drafts and highlight the importance of answer-loss allocation (Table~\ref{tab:redist-ablation}).

Our contributions are threefold:
\begin{itemize}
    \item \textbf{Formulation and evaluation principle.} We formulate incremental trace teaching utility relative to matched answer-only supervision for a specified student and evaluation distribution, distinguishing this quantity from complete-recipe performance comparisons.

    \item \textbf{Student-conditioned trace construction.} We introduce \ReDiST, which combines evidence-guided revision and repair of student drafts with selection by a frozen student-matched reader and explicit control of repair exposure and answer supervision.

    \item \textbf{Empirical evidence.} We demonstrate gains over answer-only and original-trace supervision across four target students and two model families. Component comparisons examine how trace construction and answer-loss allocation affect source-task and transfer performance.
\end{itemize}

\section{Problem Setting and Trace Teaching Utility}
\label{sec:trace-setting}

\subsection{Evidence-Rich Reasoning Distillation}
\label{sec:trace-setting-task}

We consider reasoning tasks with training data $\mathcal D=\{(Q_i,E_i,Y_i)\}_{i=1}^{n}$, where $Q_i$ is a question, $E_i$ is supplied evidence, and $Y_i$ is a verified answer. The evidence may contain both supporting information and distractors. Given an initial student $s_0$, our goal is to \textit{construct reasoning supervision that improves its learning beyond final-answer labels alone.} Construction uses these training examples and generated attempts, without requiring gold reasoning traces. At inference, the trained student answers new questions from supplied evidence.

We distinguish a generated attempt $(R,\widehat Y)$ from a supervision target $(R^\star,Y)$, where $R$ is the generated reasoning, $\widehat Y$ is the generated final answer, and $R^\star$ is the constructed trace paired with the verified answer $Y$. On a fixed training cohort, a construction $c$ specifies the targets $R_i^\star$, allowing an empty trace for answer-only supervision. A recipe $\pi$ specifies example exposure, loss allocation, and the update budget. This separates which traces are used from how they enter training.

\subsection{Student-Relative Teaching Utility}
\label{sec:trace-utility}

Let $M_{\mathcal P}(s)$ denote a higher-is-better performance metric for student $s$ on evaluation distribution $\mathcal P$ under a fixed evaluation protocol. Write $\operatorname{Train}_{s_0}(c;\pi)$ for training from $s_0$ with construction $c$ and recipe $\pi$. For a given $c$, its paired answer-only construction $c_{\varnothing}$ removes each trace while retaining the same examples and verified answers. We define incremental trace teaching utility as
\begin{equation}
\operatorname{UTIL}^{M}_{s_0,\mathcal P}(c;\pi)=M_{\mathcal P}\!\left(\operatorname{Train}_{s_0}(c;\pi)\right)-M_{\mathcal P}\!\left(\operatorname{Train}_{s_0}(c_{\varnothing};\pi)\right).
\label{eq:utility}
\end{equation}

The paired runs retain the same question IDs, evidence, answer labels, initialization, optimizer settings, update budget, and realized per-example sampling schedule, as well as the same input and output templates apart from removing the trace content. In particular, removing a trace does not change its example's exposure. The same loss rule applies to the resulting completions, although their supervised-token counts can differ, and  matching token budgets is a separate control. This quantity measures the effect of including the constructed traces under the specified training protocol. Complete-recipe comparisons may also change the cohort, exposure, or loss allocation and are interpreted separately. Evaluating Equation~\eqref{eq:utility} on the source distribution and on other benchmarks yields source-task and cross-benchmark utility, respectively. A construction can therefore have different teaching utility across evaluation distributions.

\subsection{Design Considerations: Content and Accessibility}
\label{sec:trace-design-considerations}

\textbf{Process content.}
Process content concerns the supporting facts and intermediate relations expressed in a trace beyond the final-answer label. These elements may already appear in the supplied evidence, and the trace can make their relevance and connections explicit. A conclusion-only trace omits those connections, while a more elaborate trace can connect a factual relation to the wrong intermediate entity. Assessing content therefore requires examining both the supporting statements and their connection to the question. Figure~\ref{fig:bridge-content-example} illustrates how a trace can
identify relevant entities while introducing an unsupported relation.

\textbf{Student accessibility.}
Accessibility concerns how readily the target student can use and learn from the expressed content. Wording, organization, and the granularity of intermediate steps can affect this compatibility. Familiar expression can preserve an incorrect step, whereas a valid derivation can remain difficult for a particular student to learn. 

Content and accessibility are distinct considerations whose effects can interact, and neither alone determines downstream teaching utility. They therefore need to be considered together when constructing teaching targets. This motivates the framework in Section~\ref{sec:method}, which combines evidence-guided editing of the supporting derivation with candidate selection informed by the target student's response.

\section{The \ReDiST Framework}
\label{sec:method}

\ReDiST treats trace distillation as a student-specific teaching problem: the target student shapes both the starting drafts and the selection of teaching targets.
The framework has three stages, illustrated in Figure~\ref{fig:redist-overview}: (1) evidence-guided revision or answer-assisted repair of student drafts, (2) screening and ranking by a frozen student-matched reader, and (3) distillation, where repair multiplicity $m$ controls example exposure and answer-loss weight $\alpha$ controls answer emphasis. Algorithm~\ref{alg:redist} in Appendix~\ref{app:implementation} summarizes this offline training pipeline. At inference, only the distilled student is used.

\subsection{Student-Draft Revision and Repair}
\label{sec:gred-construction}

\textbf{Draft selection.}
For each training example $(Q,E,Y)$, the initial student $s_0$ generates $K_d$ attempts $(R^{(k)},\widehat Y^{(k)})$, $k=1,\ldots,K_d$. We use a frozen student-matched reader to assign each eligible draft a verified-answer score $q_{s_0}(Y\mid Q,E,R)$, defined in Section~\ref{sec:redist-selection}. Among complete, eligible drafts, we first consider those whose generated final answer matches $Y$ and choose the one with the largest score. If none answers correctly, we choose the highest-scoring eligible draft. Denote the selected draft by $(R^{(0)},\widehat Y^{(0)})$. Cases with no reader-eligible draft, along with other parsing exceptions and format-only exclusions, follow the deterministic rules in Appendix~\ref{app:implementation}.

\textbf{Evidence-guided revision.}
When $\widehat Y^{(0)}$ matches $Y$, the rewrite model (teacher) $g$ receives $(Q,E,R^{(0)})$ and generates $K_r$ candidate revisions. It is instructed to retain supported reasoning, remove irrelevant or unsupported content, and make intermediate entities and relations explicit. The verified answer is not supplied as a separate field in this branch. The teacher may change individual steps, their ordering, and their expression to improve the derivation.

\textbf{Answer-assisted repair.}
When the answer of the selected draft does not match $Y$, the teacher additionally receives $Y$ as a repair target and generates $K_r$ candidates. It is instructed to correct the supporting derivation against $E$ and connect the relevant facts to the verified answer. This branch makes unsuccessful attempts available for trace supervision beyond correctness-only filtering. Invalid edited outputs are removed by the deterministic rules in Appendix~\ref{app:implementation}, and each candidate retains its origin as an original draft, a revision, or a repair.

\subsection{Frozen-Student Target Selection}
\label{sec:redist-selection}

An edit can change both the derivation and the student's response to it. We therefore compare edited candidates with the selected draft before assigning a teaching target. The reader remains frozen throughout construction and distillation. Given $(Q,E,R)$, it produces an answer prediction $\widehat Y^{\mathrm{read}}_{s_0}(Q,E,R)$ and a verified-answer score $q_{s_0}(Y\mid Q,E,R)$. We screen candidates by answer recovery and rank those that pass by their verified-answer score. The task-specific reader interface and scoring rules are given in Appendix~\ref{app:implementation}.

Let $\mathcal C$ contain the selected draft and edited candidates that satisfy the validity rules for reader input, with each candidate's origin retained. Define the candidates that pass the recovery check as
\begin{equation}
\mathcal C^{+} =\left\{R\in\mathcal C: \widehat Y^{\mathrm{read}}_{s_0}(Q,E,R)\text{ matches }Y\right\}.
\label{eq:redist-recovery}
\end{equation}
We select the supervision trace according to
\begin{equation}
R^\star=
\begin{cases}
\displaystyle\operatorname*{arg\,max}_{R\in\mathcal C^{+}}q_{s_0}(Y\mid Q,E,R), & \mathcal C^{+}\ne\varnothing,\\
R^{(0)}, & \mathcal C^{+}=\varnothing \text{ and }\widehat Y^{(0)}\text{ matches }Y,\\
\varnothing, & \text{otherwise}.
\end{cases}
\label{eq:redist-selection}
\end{equation}
Candidates retain their identities and origins even when their trace texts coincide, while score ties follow a fixed candidate order. If no candidate passes, we retain an initially answer-correct draft or use an empty trace for answer-only supervision. Every retained example remains paired with the verified final answer $Y$.

The reader's recovery provides a student-conditioned selection signal. Because the reader also receives the question and evidence, successful recovery does not by itself establish stepwise correctness or learning benefit.

\subsection{Distillation and Supervision Allocation}
\label{sec:redist-training}

After selecting teaching targets, we control how often accepted repairs enter training and how strongly final-answer tokens are supervised.

\textbf{Repair exposure.}
Each accepted repair appears $m$ times in the training mixture, and every other training example, including answer-only fallbacks, appears once. During training, we sample mini-batches from the training mixture.

\textbf{Answer supervision.}
Each supervised completion contains the selected trace $R_i^\star$ followed by the verified answer $Y_i$. Under a token-averaged completion loss, a long trace contributes many more supervised positions than a short answer. To make the emphasis on final-answer prediction explicit, we combine the full-completion loss with an auxiliary answer-token loss.

For a training batch, let $\mathcal T$ contain all supervised completion positions and let $\mathcal A\subseteq\mathcal T$ contain the answer-content positions. For trainable student parameters $\theta$, let $\ell_t(\theta)$ denote the next-token cross-entropy at position $t$, conditioned on the question, evidence, and preceding completion tokens. We define $\mathcal L^{\mathrm{full}}=\frac{1}{|\mathcal T|}\sum_{t\in\mathcal T}\ell_t(\theta)$ and $\mathcal L^{\mathrm{ans}}=\frac{1}{|\mathcal A|}\sum_{t\in\mathcal A}\ell_t(\theta)$.
The distillation objective is
\begin{equation}
\mathcal L =(1-\alpha)\mathcal L^{\mathrm{full}}+\alpha\mathcal L^{\mathrm{ans}}, \qquad 0\leq\alpha\leq1.
\label{eq:redist-loss}
\end{equation}
The full-completion loss already includes answer tokens, so $\alpha$ controls their additional emphasis. At $\alpha=1$, only answer positions receive direct token supervision, but their predictions still condition on the preceding trace. This therefore differs from the trace-removed answer-only control in Equation~\eqref{eq:utility}.

\textbf{Connection to content and accessibility.}
Draft-conditioned editing and frozen-student selection jointly address the content and accessibility considerations in Section~\ref{sec:trace-setting}. Repair multiplicity $m$ and answer-loss weight $\alpha$ control how the selected supervision enters training.

\section{Experiments and Results}
\label{sec:experiments}
\vspace{-0.05in}

We evaluate whether \ReDiST turns student-generated attempts into supervision that improves both source-task performance and cross-benchmark transfer. Our experiments address three questions: (i) how \ReDiST compares with answer-only, original-trace, and existing trace-selection and rewriting methods, (ii) what trace construction and answer-loss allocation contribute to the resulting performance, and (iii) whether the gains extend across target students and how the choice of rewrite teacher affects downstream performance.

\subsection{Experimental setup}
\label{sec:experimental-setup}

\textbf{Tasks and datasets.}
We use MuSiQue \citep{musique} as the primary source benchmark for training and source-task evaluation. We evaluate cross-benchmark performance on 2Wiki \citep{twowiki} and HotpotQA \citep{hotpotqa}, which provide different multi-hop questions and evidence distributions, and on QuALITY \citep{qualitydataset}, a long-context multiple-choice reading-comprehension benchmark. We report exact match (EM) and token-level F1 for MuSiQue, 2Wiki, and HotpotQA, and accuracy and hard-subset accuracy for QuALITY.

\textbf{Baselines.}
\emph{Raw Model} evaluates the student before task-specific training, \emph{Answer-only} supervises the verified final answer, and \emph{Original Trace} distills unmodified answer-correct student traces.
We also compare with STaR-style rationalization \citep{star}, student-side trace selection using NLL, RSR \citep{rsr}, and LARK \citep{lark}, and metacognitive rewriting with MBT-R \citep{mbt}. These comparisons assess the complete recipe against answer supervision, unmodified traces, trace selection, and trace rewriting. The implementation details are reported in Appendix~\ref{app:result-protocols}.

\textbf{Models and training.}
Our primary experiments use Qwen3-8B \citep{qwen3} as the target student and MuSiQue as the supervision source. Cross-student experiments additionally evaluate Qwen3-4B, Gemma4-E2B, and Gemma4-E4B \citep{gemma4}, with additional results provided in Table~\ref{tab:cross-setting-utility}. \ReDiST performs trace construction and target selection offline, and inference uses only the distilled student. Full construction and optimization details are provided in Appendix~\ref{app:implementation}.

\textbf{Evaluation protocol.}
The primary and cross-student comparisons use MuSiQue supervision, and a separate Qwen3-8B comparison uses 2Wiki as the training source (Table~\ref{tab:2wiki-transfer} in Appendix~\ref{app:additional-results}).
All \ReDiST results use $\alpha=0.15$ and $K_d=K_r=4$ unless noted otherwise. Table~\ref{tab:alpha-sweep} in Appendix~\ref{app:additional-results} reports results for other values of $\alpha$. The primary comparison uses 1,000-example evaluation slices. Table~\ref{tab:qwen8b-both-directions} in Appendix~\ref{app:additional-results} reports 95\% bootstrap intervals over the evaluation slice for all Qwen3-8B point estimates .

\begin{table}[t]
\centering
\caption{\textbf{Source-task and cross-benchmark performance after MuSiQue supervision.} All systems use Qwen3-8B. Trained variants use only MuSiQue examples for gradient updates.}
\label{tab:musique-transfer}
\small
\setlength{\tabcolsep}{8pt}
\begin{tabular}{lcccccccc}
\toprule
& \multicolumn{2}{c}{\shortstack{\textbf{MuSiQue}\\(Source)}}
& \multicolumn{2}{c}{\shortstack{\textbf{2Wiki}\\(Transfer)}}
& \multicolumn{2}{c}{\shortstack{\textbf{HotpotQA}\\(Transfer)}}
& \multicolumn{2}{c}{\shortstack{\textbf{QuALITY}\\(Transfer)}}\\
\cmidrule(lr){2-3}\cmidrule(lr){4-5}\cmidrule(lr){6-7}\cmidrule(lr){8-9}
\textbf{Method} & \textbf{EM}$\uparrow$ & \textbf{F1}$\uparrow$ & \textbf{EM}$\uparrow$ & \textbf{F1}$\uparrow$ & \textbf{EM}$\uparrow$ & \textbf{F1}$\uparrow$ & \textbf{Acc.}$\uparrow$ & \textbf{H.Acc.}$\uparrow$ \\
\midrule
Raw Model & 0.342 & 0.450 & 0.516 & 0.605 & 0.503 & 0.677 & 0.623 & 0.561\\
Answer-only & 0.562 & 0.662 & 0.479 & 0.560 & 0.497 & 0.659 & 0.523 & 0.427 \\
\midrule
Original Trace & 0.426 & 0.518 & 0.521 & 0.608 & 0.512 & 0.667 & 0.627 & 0.535\\
STaR & 0.510 & 0.613 & 0.613 & 0.689 & 0.583 & 0.745 & 0.587 & 0.541\\
NLL & 0.546 & 0.642 & 0.600 & 0.693 & 0.597 & 0.767 & 0.617 & 0.554\\
RSR & 0.568 & 0.668 & 0.624 & 0.720 & 0.600 & 0.765 & 0.577 & 0.490 \\
LARK& 0.539 & 0.626 & 0.630 & 0.712 & 0.607 & 0.774 & 0.603 & 0.522 \\
MBT-R & 0.579 & 0.672 & 0.633 & 0.719 & 0.627 & 0.768 & 0.620 & 0.567 \\
\midrule
\textbf{\ReDiST (Ours)} & \textbf{0.584} & \textbf{0.681} & \textbf{0.686} & \textbf{0.780} & \textbf{0.633} & \textbf{0.795} & \textbf{0.640} & \textbf{0.573} \\
\bottomrule
\end{tabular}%
\vspace{-0.05in}
\end{table}

\subsection{Source-task performance and transfer}
\label{sec:main-performance}

Table~\ref{tab:musique-transfer} compares Qwen3-8B models trained with MuSiQue supervision and evaluated on MuSiQue, 2Wiki, HotpotQA, and QuALITY. All gradient updates use MuSiQue examples. \ReDiST achieves the highest reported scores across all eight metrics, combining strong source-task performance with improvements over the evaluated supervision baselines on all three transfer benchmarks. 

\textbf{Source-task gains do not guarantee cross-benchmark gains.}
The answer-only and original-trace recipes illustrate this distinction. Answer-only training raises MuSiQue EM from 0.342 to 0.562, while 2Wiki EM falls from 0.516 to 0.479. Original-trace supervision reaches 0.426 and 0.521, respectively, reversing the ranking of these two recipes across benchmarks. Answer-only supervision directly trains prediction of the verified answer from the question and evidence, whereas trace supervision additionally makes intermediate statements and relations explicit learning targets. Such targets may convey evidence-linking structure that is useful on new questions, but their presence alone does not ensure effective distillation, as the weaker source-task performance of original traces illustrates. \ReDiST improves both source and transfer performance over these references.

QuALITY provides a complementary test of transfer to long-document, multiple-choice comprehension. Whereas most trace-based recipes improve 2Wiki and HotpotQA EM over the raw model, most do not improve its QuALITY performance. The raw model already reaches 0.623 accuracy and 0.561 hard-subset accuracy, while answer-only training reduces these scores to 0.523 and 0.427, respectively. \ReDiST reaches 0.640 and 0.573, modestly improving both raw-model references. Its gains on multi-hop benchmarks are therefore accompanied by preservation and modest improvement of performance on this more distinct transfer task. This pattern highlights the importance of evaluating both gains on related benchmarks and possible degradation on other question-answering tasks.

\textbf{Similar source accuracy, stronger 2Wiki transfer.}
Among the trace-construction baselines, MBT-R is the closest comparison because it also rewrites student attempts and uses answer guidance to correct unsuccessful reasoning. MBT-R and \ReDiST achieve similar MuSiQue EM (0.579 vs. 0.584), but differ substantially on 2Wiki (0.633 vs. 0.686), a 5.3-point gap. The difference is smaller on HotpotQA (0.627 vs. 0.633), showing that the transfer advantage depends on the benchmark. Table~\ref{tab:2wiki-transfer} in Appendix~\ref{app:additional-results} extends this comparison to Qwen3-8B trained with 2Wiki supervision. \ReDiST reaches 0.727 source EM and 0.536 MuSiQue transfer EM, compared with 0.707 and 0.512 for MBT-R. It also achieves 0.660 QuALITY accuracy versus 0.623 for MBT-R. These results extend the source-task and cross-benchmark gains beyond MuSiQue supervision, including the reverse 2Wiki-to-MuSiQue transfer direction.

\begin{figure}[t]
\centering
\includegraphics[width=\textwidth]{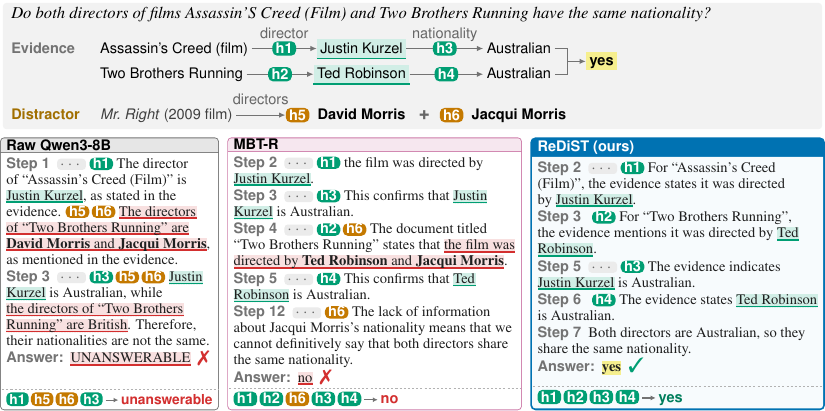}
\vspace{-0.1in}
\caption{\textbf{Evidence use in a selected 2Wiki example.} Test-time reasoning excerpts from raw Qwen3-8B and Qwen3-8B students trained with MBT-R or \ReDiST. MBT-R identifies the relevant directors but introduces an unsupported co-director. In contrast, the \ReDiST output uses the relevant film--director--nationality links and answers correctly. Quoted model text is reproduced verbatim, with ellipses marking omissions, while highlights, correctness markers, and notes are author annotations.}
\label{fig:bridge-content-example}
\vspace{-0.1in}
\end{figure}

\textbf{Relevant facts can coexist with unsupported relations.}
In the selected 2Wiki example in Figure~\ref{fig:bridge-content-example}, MBT-R identifies the relevant directors and their nationalities but adds a co-director from a distractor film. Its trace then treats that person's unestablished nationality as relevant and answers incorrectly. In contrast, the \ReDiST-trained student restricts the comparison to the two queried directors and answers correctly. \textit{This case illustrates why process assessment should check unsupported additions alongside the required evidence links.
}

These quantitative results compare complete training recipes, whereas the following component comparisons further examine the contributions of trace construction and answer-loss allocation.

\subsection{Trace Construction and Supervision Allocation}
\label{sec:ablation}
Table~\ref{tab:redist-ablation} examines two decisions in \ReDiST: which intermediate targets to distill and how to allocate supervision to those targets. We first compare alternative trace constructions, then examine answer-loss allocation with the selected targets held fixed. Appendix~\ref{app:ablation-matching} details the controls used for these comparisons.

\begin{table}[t]
\centering
\caption{\textbf{\ReDiST component and supervision-allocation ablations.} Variants modify trace construction or answer-loss allocation relative to the full recipe. All entries are EM.}
\label{tab:redist-ablation}
\small
\setlength{\tabcolsep}{4pt}
\begin{tabular}{llccc}
\toprule
& & Source & \multicolumn{2}{c}{Transfer} \\
\cmidrule(lr){3-3}\cmidrule(lr){4-5}
Variant & Controlled change & MuSiQue & 2Wiki & HotpotQA \\
\midrule
Full \ReDiST & Reference recipe & 0.584 & \textbf{0.686} & \textbf{0.633} \\
\textsc{Original target} & No revision or repair & 0.426 & 0.521 & 0.512 \\
\textsc{Selected student target} & No revision or repair & 0.552 & 0.577 & 0.550 \\
\textsc{Fresh rewrite} & Rewrite without the original draft & 0.554 & 0.483 & 0.570 \\
\textsc{Repair disabled} & Original targets for repair cases & 0.577 & 0.626 & 0.576 \\
\textsc{No answer auxiliary} & $\alpha=0$ in Eq.~\eqref{eq:redist-loss} & \textbf{0.592} & 0.652 & 0.630 \\
\textsc{Answer-loss only} & $\alpha=1$ in Eq.~\eqref{eq:redist-loss} & 0.573 & 0.417 & 0.480 \\
\textsc{Trace-only control} & Mask answer positions in loss & 0.404 & 0.478 & 0.443 \\
\bottomrule
\end{tabular}%
\vspace{-0.1in}
\end{table}

\textbf{Construction adds gains beyond selecting student drafts.}
Compared with \textsc{Original target}, \textsc{Selected student target} raises MuSiQue EM from 0.426 to 0.552 (+12.6 points) and also improves both transfer benchmarks. Full \ReDiST improves further over \textsc{Selected student target}: MuSiQue EM rises from 0.552 to 0.584 (+3.2 points), while 2Wiki EM rises from 0.577 to 0.686 (+10.9 points) and HotpotQA EM from 0.550 to 0.633 (+8.3 points). Thus, \textit{choosing among existing attempts already improves supervision, while the full construction pipeline yields additional gains that are larger on the transfer benchmarks.} This supports using the student's attempts as starting points for constructing teaching targets beyond those it initially generates.

\textbf{Draft conditioning and repair matter most for transfer.}
Compared with full \ReDiST, \textsc{Fresh rewrite} omits the student draft from rewrite input while retaining the same evidence and branch-specific answer access. MuSiQue EM falls from 0.584 to 0.554, 2Wiki EM from 0.686 to 0.483, and HotpotQA EM from 0.633 to 0.570. The largest decline is 20.3 points on 2Wiki. Fresh rewriting also falls below \textsc{Selected student target} on 2Wiki (0.483 vs. 0.577), despite similar MuSiQue EM (0.554 vs. 0.552). \textit{Conditioning the rewrite on the student's draft therefore yields better supervision than generating a new derivation.} The advantage of draft-conditioned construction supports using the student's existing reasoning path to inform what the teacher revises, even when substantial changes are allowed.

Comparing full \ReDiST with \textsc{Repair disabled} reveals a similar transfer-oriented pattern. Disabling repair reduces MuSiQue EM from 0.584 to 0.577, while 2Wiki EM falls from 0.686 to 0.626 and HotpotQA EM from 0.633 to 0.576. The repair-disabled control retains the same questions and uses original-target fallbacks for repair cases. With question coverage preserved, this comparison supports improving the supervision attached to initially unsuccessful attempts. These effects are less visible in source-task EM, making transfer evaluation particularly informative for draft conditioning and repair.

\textbf{Answer weighting matters even when the traces are fixed.} The three allocation variants at the bottom of Table~\ref{tab:redist-ablation} train on exactly the training mixture of full \ReDiST; they differ only in which supervised positions carry loss and how they are weighted. \textsc{No Answer Auxiliary} sets $\alpha = 0$ in Equation~\eqref{eq:redist-loss}, so answer positions are weighted like any other position, and reaches 0.592 MuSiQue EM, 0.652 2Wiki EM, and 0.630 HotpotQA EM. The full recipe with $\alpha = 0.15$ is slightly lower on MuSiQue (0.584), essentially equal on HotpotQA (0.633), and 3.4 points higher on 2Wiki (0.686); Table~\ref{tab:alpha-sweep} in Appendix~\ref{app:additional-results} shows that larger values of $\alpha$ lower transfer EM. \textsc{Answer-Loss Only} takes $\alpha = 1$, so only answer positions carry loss. It retains most of the source-task performance (0.573), but transfer falls to 0.417 on 2Wiki and 0.480 on HotpotQA, below the raw model. \textsc{Trace-Only Control} masks the answer positions out of the loss, and EM falls to 0.404, 0.478, and 0.443. Because long traces already leave the answer with a small share of a uniformly averaged objective, these results indicate that both parts of the completion need loss, with the answer share small but explicit: dropping answer supervision is harmful, supervising the answer alone forfeits the transfer gains that trace supervision provides, uniform averaging matches or slightly exceeds the full recipe on the source task, and moderate reweighting mainly benefits transfer to 2Wiki. Trace construction determines the intermediate supervision available to the student, while answer weighting controls the emphasis placed on predicting the verified final answer. Together, these comparisons support evaluating both the construction of intermediate targets and the training objective used to learn from them, with the largest gains over answer supervision appearing on the transfer benchmarks.

\vspace{-0.05in}
\subsection{Generality across Student and Teacher Models}
\label{sec:generality}
\vspace{-0.05in}
We evaluate \ReDiST across four target students and conduct a separate sweep of rewrite teachers. These experiments examine whether its benefits extend beyond the primary student and how teacher choice affects the resulting supervision. All settings use MuSiQue as the source of training examples.

\begin{table}[t!]
\centering
\caption{\textbf{Source-task and transfer performance across target students.} All trained variants receive MuSiQue supervision. Columns report MuSiQue source-task (S) and 2Wiki transfer (T) EM for each student.}
\label{tab:cross-setting-utility}
\small
\setlength{\tabcolsep}{8pt}
\begin{tabular}{lcccccccc}
\toprule
& \multicolumn{2}{c}{Qwen3-4B} & \multicolumn{2}{c}{Qwen3-8B} & \multicolumn{2}{c}{Gemma4-E2B} & \multicolumn{2}{c}{Gemma4-E4B} \\
\cmidrule(lr){2-3}\cmidrule(lr){4-5}\cmidrule(lr){6-7}\cmidrule(lr){8-9}
Supervision & S & T & S & T & S & T & S & T \\
\midrule
Raw model & 0.293 & 0.225 & 0.342 & 0.516 & 0.225 & 0.518 & 0.425 & 0.643 \\
Answer-only & 0.558 & 0.440 & 0.562 & 0.479 & 0.514 & 0.368 & 0.578 & 0.449 \\
Original trace & 0.405 & 0.445 & 0.426 & 0.521 & 0.311 & 0.546 & 0.456 & 0.644 \\
STaR & 0.519 & 0.480 & 0.510 & 0.613 & 0.444 & 0.588 & 0.553 & 0.682 \\
NLL & 0.519 & \textbf{0.635} & 0.546 & 0.600 & 0.545 & 0.610 & 0.571 & 0.682 \\
RSR & 0.533 & 0.587 & 0.568 & 0.624 & 0.542 & 0.607 & 0.551 & 0.680 \\
LARK & 0.507 & 0.626 & 0.539 & 0.630 & 0.539 & 0.600 & 0.559 & 0.665 \\
MBT-R & 0.556 & 0.625 & 0.579 & 0.633 & 0.542 & 0.580 & 0.606 & 0.645 \\
\textbf{\ReDiST (Ours)} & \textbf{0.580} & 0.630 & \textbf{0.584} & \textbf{0.686} & \textbf{0.559} & \textbf{0.622} & \textbf{0.608} & \textbf{0.723} \\
\bottomrule
\end{tabular}
\vspace{-0.1in}
\end{table}

\textbf{Gains extend across target students.}
Table~\ref{tab:cross-setting-utility} compares four students from the Qwen3 and Gemma4 families. \ReDiST exceeds answer-only and original-trace supervision on both benchmarks for every student. It achieves the highest reported MuSiQue EM in all four comparisons and the highest 2Wiki EM in three. Beyond the primary Qwen3-8B comparison, Gemma4-E4B reaches 0.723 on 2Wiki, compared with 0.682 for the strongest alternatives, STaR and NLL, while Gemma4-E2B reaches 0.622, compared with 0.610 for NLL. Qwen3-4B is the exception: \ReDiST reaches 0.630, slightly below NLL at 0.635. Appendix~\ref{app:additional-results} repeats this comparison with 2Wiki supervision (Table~\ref{tab:cross-setting-utility-2wiki}) and reports all four benchmarks for Qwen3-4B (Table~\ref{tab:musique-transfer-qwen3-4b}). \textit{These results support the applicability of the construction and distillation pipeline across the evaluated model families and scales, while showing that its advantage over competing strategies depends on the target student.}

The source-transfer distinction also persists across students. Answer-only supervision outperforms original traces on MuSiQue but underperforms them on 2Wiki for all four students. For Gemma4-E4B, for example, answer-only supervision achieves higher source EM than original traces (0.578 vs. 0.456), but lower transfer EM (0.449 vs. 0.644). Thus, the different supervision profiles observed in Table~\ref{tab:musique-transfer} extend beyond Qwen3-8B, \textit{reinforcing the value of evaluating supervision on both source and transfer distributions when the learner changes.}

\textbf{Similar repair success can accompany different transfer outcomes.}
A separate rewrite-teacher sweep with Qwen3-8B student initialization reinforces the distinction between construction-time signals and downstream performance (Table~\ref{tab:gred-teacher-sweep} in Appendix~\ref{app:additional-results}): Gemma4-26B-A4B has higher reported repair success than Qwen3-14B (62.15\% versus 61.78\%) but yields lower 2Wiki EM (0.624 vs. 0.686). This supports evaluating rewrite teachers through the distilled student's performance on the intended evaluation distributions.

\providecommand{\blankcell}{--}

 \vspace{-0.1in}
\section{Conclusion}
\label{sec:discussion}
\vspace{-0.1in}

We introduced \ReDiST, which constructs reasoning supervision through evidence-guided revision of student drafts, answer-assisted repair, and frozen-student target selection. Under MuSiQue supervision, it improves source-task and 2Wiki performance over answer-only and original-trace supervision across four students. A separate Qwen3-8B comparison extends these gains to 2Wiki supervision and transfer to MuSiQue. Component comparisons show larger transfer gains from draft conditioning and repair, while loss ablations with the selected traces held fixed support joint supervision of reasoning traces and final answers.
These findings support designing trace targets and their supervision together for the intended learner, with teaching value assessed through both source-task and cross-benchmark performance.

\bibliographystyle{assets/plainnat}
\bibliography{mybib}

\clearpage
\beginappendix

\section{Related Work}
\label{app:related-work}

\paragraph{Trace construction and self-training.}
One line of work improves reasoning supervision by generating, correcting, or restructuring the traces used for training. STaR bootstraps reasoning from model-generated rationales and uses answer-conditioned rationalization when initial attempts fail \citep{star}. Re-ReST repairs low-quality trajectories using external feedback \citep{rerest}, and SuperCorrect and CGD incorporate correction into distillation \citep{supercorrect,cgd}. Metacognitive Behavioral Tuning rewrites or synthesizes traces with an explicit multi-stage metacognitive structure \citep{mbt}. Its MBT-R variant also rewrites student attempts and uses answer guidance to correct unsuccessful reasoning, making it our closest baseline. Controlled work shows that answer-conditioned rationalization can degrade verifiable-reasoning distillation even when the final answer is correct \citep{answerconditioned}. \ReDiST constructs teaching targets by editing the target student's own attempts and gives the verified answer only to the repair of unsuccessful drafts. Our fresh-rewrite control indicates that this draft conditioning matters most for transfer (Section~\ref{sec:ablation}).

\paragraph{Trajectory selection and learner compatibility.}
A second line of work asks which traces a particular student can learn from, and selects, scores, or adapts trajectories accordingly. RSR measures student-specific informative alignment through rank and surprisal \citep{rsr}, and LARK selects trajectories using a learnability proxy and distributional coverage \citep{lark}. The Quality--Utility Paradox shows that higher reward-model quality can be offset by learner--data mismatch and addresses this mismatch through style-preserving refinement \citep{qualityutility}. DAIL and TSD-KD address learner--teacher mismatch \citep{dail,tsdkd}. RLT and COALITION use student answer prediction to optimize explanations or select rationales \citep{rlt,coalition}, while TRM and ReasonFlux-PRM score reasoning quality \citep{trm,reasonfluxprm}. \ReDiST first creates new candidates by editing the student's own drafts and then chooses among the original and edited versions with a frozen student-matched reader, using this student signal offline without training the rewrite model.

\paragraph{Distillation objectives and training data.}
Distillation outcomes also depend on how traces enter training and on the composition of the training data. Distilling step-by-step trains rationale generation and label prediction as separate tasks \citep{hsieh2023distilling}. OPSD distills a privileged self-teacher into on-policy student rollouts, ROSD adds reflection-guided, error-localized correction, and SDPO turns textual feedback into dense self-distillation signals during reinforcement learning \citep{opsd,rosd,sdpo}. Reasoning Scaffolding distills reasoning structure \citep{reasoningscaffolding}, and influence-based attribution relates training data to reasoning gains \citep{infra}. NaturalThoughts studies the scale, difficulty, and diversity of distilled reasoning data and its mixture with answer-only data \citep{naturalthoughts}, and DC-CoT benchmarks data selection and mixing for chain-of-thought distillation and their effect on generalization \citep{dccot}. Unlike on-policy self-distillation, \ReDiST constructs targets offline and trains with supervised fine-tuning on a single completion containing the trace and the verified answer.

\paragraph{Evidence grounding and process measurement.}
Our setting also connects to work on reasoning over supplied evidence and on what visible reasoning traces reveal. In retrieval-augmented and evidence-rich question answering, prior work trains models to be robust to irrelevant context \citep{yoran2024making} and constructs knowledge-grounded reasoning chains from retrieved documents \citep{fang2024trace}. Chain-of-thought explanations need not faithfully reflect the computation behind a prediction \citep{turpin2023language,arcuschin2025chain}, and information-bottleneck and information-flow views analyze reasoning compression and faithfulness \citep{reasoningcompression,jia2026informationflow}. Predictive $\mathcal{V}$-information formalizes the information that a restricted computational family can extract \citep{vinfo}. \ReDiST uses the supplied evidence to revise and repair the student's own reasoning before distillation and treats visible traces only as teaching targets.

\section{Experimental and Implementation Details}
\label{app:extended-design}

This appendix specifies the construction, training, and evaluation details supporting the main comparisons.

\subsection{Tasks, students, and comparison endpoints}

The primary comparison trains Qwen3-8B with MuSiQue supervision and evaluates source-task performance and transfer to 2Wiki \citep{twowiki}, HotpotQA \citep{hotpotqa}, and QuALITY \citep{qualitydataset}. The model sweep varies the target student under MuSiQue supervision. A separate second-source study trains Qwen3-8B on 2Wiki and reports source-task and transfer outcomes. The rewrite-teacher sweep varies the rewrite model with a fixed Qwen3-8B student initialization. These studies separate the effects of changing the learner, the training source, and the rewrite teacher.

QA results use answer EM and token F1, and QuALITY results use accuracy and hard-subset accuracy. Each metric is interpreted within its own task. The raw model anchors performance before task-specific training, and each student/task setting has its own answer-only reference for Equation~\eqref{eq:utility}.

\subsection{\ReDiST implementation}
\label{app:implementation}
\label{sec:trace-families}

\paragraph{Draft generation and selection.}
For each training example $(Q,E,Y)$, \ReDiST first generates $K_d$ candidate drafts from the target student. The implementation uses one deterministic draft followed by sampled drafts. Unless otherwise specified, the frozen reader is a copy of the target student. A draft is reader-eligible only when it contains a complete nonempty reasoning trace and a well-formed final answer. 
For an eligible trace $R$, the reader is given the student prompt followed by $R$. Its answer prediction $\hat{Y}^{\mathrm{read}}_{s_0}(Q,E,R)$ is the greedy continuation from this position up to answer closing tag, and its verified-answer score $q_{s_0}(Y \mid Q,E,R)$ is the product of the reader's next-token probabilities over the tokens of $Y$.
\ReDiST first restricts attention to drafts whose generated final answer matches $Y$ when such drafts exist, and otherwise considers all reader-eligible drafts. In either case, it selects the candidate with the largest reader score. Ties are resolved deterministically by candidate order. If no draft is reader-eligible, the implementation falls back to the first generated draft for deterministic case classification.

\paragraph{Revision and repair.}
The selected draft determines which rewrite prompt is used. For an answer-correct draft, the rewrite model receives only the question, evidence, and draft reasoning. For an unsuccessful draft, the verified answer is additionally supplied as a checking target for answer-assisted repair. In both branches, the rewrite model is instructed to produce a complete evidence-grounded explanation, retain useful supported reasoning, remove unsupported or irrelevant content, and explicitly connect the relevant intermediate entities and relations. The rewrite is allowed to reorder, replace, add, or remove steps and to change the wording, so \ReDiST conditions on an existing draft without imposing a minimal-edit constraint. Each branch generates $K_r$ rewrite candidates. Empty or length-capped generations, traces containing reasoning/answer tags, and direct answer-declaration patterns are removed by deterministic validity filters.

\paragraph{Student-conditioned target selection.}
The original draft and all valid rewrite candidates are rescored by the frozen reader. A candidate passes the functional answer-recovery check when the reader's greedy answer $\hat{Y}^{\mathrm{read}}_{s_0}(Q,E,R)$ matches $Y$ under the normalized exact-match rule used for evaluation. If at least one candidate passes, \ReDiST selects the passing candidate with the largest $q_{s_0}(Y\mid Q,E,R)$. If none passes and the original draft was answer-correct, the original draft is retained. If none passes for an unsuccessful draft, the target becomes answer-only. A selected generated candidate from the unsuccessful branch is labeled as a repair, while a selected original or correct-branch candidate is labeled as retained or revised. The answer-recovery check provides a functional selection criterion.

\paragraph{Validity and format-only cases.}
Before rewrite and training, deterministic parsing distinguishes ordinary unsuccessful drafts from answer-like or format-only cases. The current implementation excludes these format-only cases from training. Controlled ablations define their shared cohort after this deterministic preprocessing step so that branch-specific interventions do not change the training question set.

\paragraph{Optimization.}
The \ReDiST SFT implementation uses rank-32 LoRA \citep{lora} with LoRA scaling 64 and dropout 0.05 on the attention projection modules and the MLP gate/up/down projections. Training uses AdamW with learning rate $10^{-4}$, minibatch size 2, gradient accumulation over 8 minibatches, cosine learning-rate decay with a short warmup, gradient clipping at norm 1, and gradient checkpointing. Base-model parameters are loaded in bfloat16 and frozen outside the LoRA adapters. All runs use 500 optimizer steps with the same batch size. Each mini-batch is drawn from the method's own training mixture. Except for MBT-R, we use a generation limit of 1,024 tokens. We use seed 42 to randomly sample 4,000 training examples from the full training set. Unless otherwise specified, we set $K_d=K_r=4$, $m=4$, and $\alpha=0.15$. Unless noted otherwise, the rewrite model is Qwen3-14B for Qwen students and Gemma4-26B-A4B for Gemma students, and the frozen reader is a copy of the target student.

\paragraph{Algorithm summary.}
In Algorithm~\ref{alg:redist}, \textsc{SelectDraft} follows the draft-selection rules in Section~\ref{sec:gred-construction} and returns $\bot$ only for the documented format-only exclusions. \textsc{BuildCandidates} applies the validity rules and preserves candidate origins. \textsc{SelectRecoverable} uses the frozen reader to apply Equation~\eqref{eq:redist-selection} and returns the selected trace $R^\star$ with its origin $\tau^\star$, including the original-draft and answer-only fallbacks.

\begin{algorithm}[t]
\caption{\ReDiST trace construction and distillation}
\label{alg:redist}
\begin{algorithmic}[1]
\Require Training data $\mathcal D$; initial student $s_0$; teacher $g$; frozen student-matched reader; draft/edit counts $K_d,K_r$; repair multiplicity $m$; answer-loss weight $\alpha$
\Ensure Distilled student $s^\star$
\State $\widetilde{\mathcal D}\gets\emptyset$
\ForAll{$(Q,E,Y)\in\mathcal D$}
    \State $\mathcal A_{\mathrm{draft}}\gets
        \Call{GenerateAttempts}{s_0,Q,E,K_d}$
    \State $a_0\gets
        \Call{SelectDraft}{\mathcal A_{\mathrm{draft}},Q,E,Y,q_{s_0}}$
    \If{$a_0=\bot$}
        \State \textbf{continue}
        \Comment{Documented parsing exclusions}
    \EndIf
    \State $(R^{(0)},\widehat Y^{(0)})\gets a_0$
    \State $b\gets[\widehat Y^{(0)}\text{ matches }Y]$
    \If{$b$}
        \State $\mathcal R_{\mathrm{new}}\gets
            \Call{Revise}{g,Q,E,R^{(0)},K_r}$
    \Else
        \State $\mathcal R_{\mathrm{new}}\gets
            \Call{Repair}{g,Q,E,R^{(0)},Y,K_r}$
    \EndIf
    \State $\mathcal C\gets
        \Call{BuildCandidates}{R^{(0)},\mathcal R_{\mathrm{new}},b}$
    \State $(R^\star,\tau^\star)\gets
        \Call{SelectRecoverable}{\mathcal C,a_0,Q,E,Y}$
    \State $w\gets m$ if $\tau^\star=\textsc{Repair}$, else $w\gets1$
    \State Append $w$ copies of $(Q,E,R^\star,Y)$ to $\widetilde{\mathcal D}$
\EndFor
\State $s^\star\gets
    \Call{Distill}{s_0,\widetilde{\mathcal D},\alpha}$ using Equation~\eqref{eq:redist-loss}
\State \Return $s^\star$
\end{algorithmic}
\end{algorithm}

\subsection{Baseline implementations}
\label{app:result-protocols}

All baselines use the same target student, the same 4{,}000 training questions, the same LoRA and optimizer configuration, the same number of optimizer steps, and the same evaluation setting as \ReDiST. They differ mainly in the supervision target attached to each training question, and therefore in which questions are retained. We group the baselines by where the supervision target comes from (Table~\ref{tab:baseline-taxonomy}): a \emph{reference} without task training, \emph{answer supervision}, traces \emph{self-generated} by the student, traces \emph{selected} from a teacher candidate pool by a student-side score, and traces \emph{rewritten} by a teacher.

\begin{table}[h]
\centering
\small
\caption{\textbf{Baseline taxonomy.} ``Student'' denotes the untrained target
student. The teacher is Qwen3-14B for Qwen students and Gemma4-26B-A4B for
Gemma students.}
\label{tab:baseline-taxonomy}
\setlength{\tabcolsep}{3pt}
\resizebox{\linewidth}{!}{%
\begin{tabular}{llllc}
\toprule
Category & Baseline & Trace source & Acceptance / selection & \\
\midrule
Reference            & Raw Model      & --- (no training)          & ---                                \\
Answer supervision   & Answer-only    & --- (gold answer only)     & aligned question set                 \\
\midrule
Self-generated       & Original Trace & student, greedy            & EM match, non-empty trace             & \\
                     & STaR           & student, greedy (+ hint)   & EM match, non-empty trace, closed tag (hint removed for training) & \\
\midrule
Trace selection      & NLL            & teacher, multi-prompt pool & $\arg\min$ mean student NLL            \\
                     & RSR            & teacher, multi-prompt pool & $\arg\min$ rank--surprisal ratio       \\
                     & LARK           & teacher, multi-prompt pool & $\arg\max$ LARK score         \\
\midrule
Trace rewriting      & MBT-R          & teacher rewrite of student draft & none (one rewrite per question) \\
\bottomrule
\end{tabular}}
\end{table}

\paragraph{Shared protocol.}
The training source is a fixed set of 4{,}000 questions drawn uniformly without replacement (seed 42) from the training split. Every question keeps its full evidence set, with supporting and distractor paragraphs and no truncation. All methods start from these 4{,}000 questions. Original Trace and STaR retain questions according to their own acceptance rules. The trace-selection arms retain the questions that have at least one usable teacher candidate, and Answer-only trains on all 4,000 questions; MBT-R rewrites every parseable draft under MuSiQue supervision and is restricted to the selection cohort under 2Wiki supervision. Every table entry evaluates the final checkpoint of its run on the first 1{,}000 examples of the validation split, and QuALITY entries use a fixed 300-question set.

\paragraph{Output format.}
Trace-supervised targets place the reasoning in a \texttt{<think>}\,\ldots\,\texttt{</think>} block, followed by \texttt{<answer>}\,gold answer\,\texttt{</answer>} and the end-of-sequence token. Qwen students are prompted with a raw-text instruction that ends inside an open \texttt{<think>} block, so their native chat-template thinking mode is not used. Gemma students use their chat template with thinking disabled. The answer span of every training target carries the dataset's gold string, regardless of the answer written in the trace. Every Qwen prompt, including the raw, answer-only, and trace prompts, also allows the answer \emph{UNANSWERABLE} when the evidence is insufficient, and such answers are scored as incorrect. The Gemma prompts have no abstention option, so Raw and Answer-only values should be compared within a model family.

\paragraph{Evaluation.}
Decoding is greedy. Exact match (EM) and token-level F1 are computed after standard answer normalization, taking the maximum over the gold answer and its annotated aliases, over all 1{,}000 examples of each benchmark. Tagged outputs are scored on the content of the first closed \texttt{<answer>}\,\ldots\,\texttt{</answer>} span. Outputs without a closed answer span are scored on their first non-empty line. For Qwen students, this line is the opening \texttt{<think>} tag and scores zero. For Gemma students, it is the first generated line, which occasionally earns partial F1. Answer-only outputs are scored on their first non-empty line. The generation budget is 1,024 new tokens by default.

\paragraph{Methods with self-generated traces.}
Following STaR~\citep{star}, questions the student already solves keep their native trace. For every rejected question, the same untrained student generates a rationalization with the gold answer appended to the question as a hint (``\emph{(The correct answer is: \ldots)}''), using greedy decoding with a 768-token budget. A rationalization is accepted only if its own final answer matches the gold answer under normalized exact match, its reasoning block is non-empty, and its answer tag is closed. Questions that fail both passes are discarded. The hint is removed from the training prompt, and accepted rationalizations are otherwise used unchanged, so some of them refer to the provided answer.

\paragraph{Methods with teacher trace selection.}
NLL, RSR, and LARK select from one shared pool generated by a teacher from the student's own model family: Qwen3-14B for Qwen students and Gemma4-26B-A4B for Gemma students. The teacher answers every training question under several gold-blind prompt variants that differ only in the requested reasoning style. We use six variants (standard, decomposition, evidence chain, entity tracking, verification, minimal chain), each replacing the reasoning line of the standard prompt. Within each question, exact duplicate traces are collapsed to a single candidate. Each method follows the selection rules given in Table~\ref{tab:baseline-taxonomy}.

\paragraph{Trace rewriting methods.}
We follow the two-turn rewriting protocol of MBT-R~\citep{mbt} with its official prompt templates, using Qwen3-14B for Qwen students and Gemma4-26B-A4B for Gemma students as the rewriting teacher, so that all teacher-generated schemes share the same teacher.

\paragraph{Retained training data.}
Table~\ref{tab:training-mixtures} lists how many of the 4,000 sampled questions each method retains and what its training mixture contains. Answer-only keeps every question; Original trace keeps only answer-correct greedy drafts; STaR recovers rejected questions through hinted rationalization, and five additional examples were excluded due to malformed or unparsable initial outputs before the rationalization pass; the trace-selection arms keep questions with at least one usable teacher candidate; MBT-R rewrites every draft, except questions on which the teacher's rewrite falls into a repetition loop and exhausts the generation budget; and \ReDiST{} drops only the questions whose selected draft violates the requested output format. Under MuSiQue supervision, 1,004 of 1,625 unsuccessful drafts are repaired and 621 fall back to answer-only targets; under 2Wiki supervision, 1,089 of 1,371 are successfully repaired.

\begin{table}[h]
\centering
\caption{\textbf{Training data retained by each supervision method (Qwen3-8B student).}
Kept counts distinct training questions.}
\label{tab:training-mixtures}
\small
\setlength{\tabcolsep}{4pt}
\begin{tabular}{l r r p{6.4cm} r}
\toprule
Method & Kept & Dropped & Training mixture & \\
\midrule
\multicolumn{5}{@{}l}{\textit{MuSiQue supervision (4,000 training questions)}} \\
Answer-only      & 4,000 &     0 & 4,000 gold answers, no trace\\
Original trace   & 1,466 & 2,534 & 1,466 answer-correct greedy traces\\
STaR             & 3,671 &   329 & 1,466 native traces $+$ 2,205 accepted rationalizations (of 2,529 attempted) \\
NLL / RSR / LARK & 2,477 & 1,523 & 2,477 selected teacher traces, one per question \\
MBT-R            & 3,995 &     5 & 3,995 rewrites of the student draft \\
\ReDiST{}        & 3,691 &   309 & 1,813 revised $+$ 253 retained correct drafts; 1004 accepted repairs; 621 answer-only fallbacks \\
\midrule
\multicolumn{5}{@{}l}{\textit{2Wiki supervision (4,000 training questions)}} \\
Answer-only      & 4,000 &     0 & 4,000 gold answers, no trace \\
Original trace   & 2,296 & 1,704 & 2,296 answer-correct greedy traces \\
STaR             & 3,895 &   105 & 2,296 native traces $+$ 1,599 accepted rationalizations (of 1,704 attempted)  \\
NLL / RSR / LARK & 3,408 &   592 & 3,408 selected teacher traces, one per question \\
MBT-R            & 3,408 &   592 & 3,408 rewrites of the student draft (selection cohort) \\
\ReDiST{}        & 3,933 &   67 & 2,439 revised $+$ 123 retained correct drafts; 1,089 accepted repairs; 282 answer-only fallbacks \\
\bottomrule
\end{tabular}
\end{table}

\subsection{Ablation details}
\label{app:ablation-matching}

Table~\ref{tab:redist-ablation} compares full \ReDiST with different variants. \textsc{Original target} is the Original Trace baseline of Table~\ref{tab:musique-transfer}, which trains on unmodified answer-correct student drafts. \textsc{Selected student target} keeps the selected draft of every question as its target, without revision or repair. \textsc{Fresh rewrite} gives the rewrite model the question, the evidence, and the same branch-specific answer access as \ReDiST, omits the student draft and its answer, and applies the same checks and fallbacks to the resulting candidates. \textsc{Repair disabled} uses the original draft as the target for every question whose selected draft was unsuccessful. These drafts can reach a wrong answer, while the supervised final answer remains the verified answer. In \textsc{Selected student target} and \textsc{Repair disabled}, a question whose draft is empty receives an answer-only target. \textsc{Selected student target}, \textsc{Fresh rewrite}, and \textsc{Repair disabled} keep the repair exposure of full \ReDiST, so questions with accepted repairs in \ReDiST appear $m$ times in training mixture even when their new targets are not repairs. \textsc{No Answer Auxiliary}, \textsc{Answer-Loss Only}, and \textsc{Trace-Only Control} all train on exactly the training mixture of full \ReDiST, with the same selected targets, the same repair exposure $m$, the same tagged prompt and output format, and the same optimizer settings; only the loss allocation over supervised positions changes. \textsc{No Answer Auxiliary} sets $\alpha = 0$ in Equation~\eqref{eq:redist-loss}, which reduces the objective to a uniform token-averaged cross-entropy over all completion positions. \textsc{Answer-Loss Only} sets $\alpha = 1$, so only answer-content positions carry loss while the trace remains in the completion as conditioning context. \textsc{Trace-Only Control} computes the loss only over trace positions and excludes the answer-content positions, so answer tokens receive no direct supervision.

\section{Additional Results}
\label{app:additional-results}

\paragraph{Rewrite-teacher sweep.}
Table~\ref{tab:gred-teacher-sweep} varies the rewrite teacher while initializing each distilled student from Qwen3-8B. MuSiQue EM ranges from 0.578 to 0.593 across teachers, whereas 2Wiki EM ranges from 0.620 to 0.686. The highest reported repair success belongs to Gemma4-26B-A4B (62.15\%), slightly above Qwen3-14B (61.78\%), but their resulting students reach 0.624 and 0.686 2Wiki EM, respectively. The Qwen3-8B teacher gives the highest source-task EM at 0.593. Thus, repair success, source-task accuracy, and transfer performance favor different rewrite teachers. Teacher choice should be evaluated through the distilled student's performance on the intended evaluation distributions.

Token-level normalized edit distance (Token NED) is the minimum number of token insertions, deletions, and substitutions between the draft and edited trace, divided by the longer trace length and expressed as a percentage. It provides a complementary description of the two editing branches. Repair NED exceeds revision NED for every teacher, and for Qwen3-14B, the values are 29.90\% and 1.15\%, respectively. This pattern is consistent with different editing roles assigned to unsuccessful and answer-correct attempts. Yet similar repair-edit magnitudes can accompany different transfer outcomes: Qwen3-14B and Gemma4-26B-A4B have repair NED values of 29.90\% and 30.04\%, despite their different 2Wiki EM scores above. Edit magnitude characterizes how much the trace changes, while downstream evaluation assesses the usefulness of the resulting supervision.

\begin{table}[h]
\centering
\caption{\textbf{Rewrite-teacher comparison with a fixed student initialization.} Each student is initialized from Qwen3-8B and distilled using MuSiQue supervision. We report repair success, token-level normalized edit distance (NED), and downstream EM/F1.}
\label{tab:gred-teacher-sweep}
\small
\setlength{\tabcolsep}{6pt}
\begin{tabular}{lccccccc}
\toprule
Rewrite teacher
& Repair success
& \multicolumn{2}{c}{Token NED (\%)}
& \multicolumn{2}{c}{MuSiQue (Source)}
& \multicolumn{2}{c}{2Wiki (Transfer)} \\
\cmidrule(lr){3-4}\cmidrule(lr){5-6}\cmidrule(lr){7-8}
& (\%) & Revise & Repair & EM & F1 & EM & F1 \\
\midrule
Qwen3-4B
& 57.97 & 0.98 & 18.31 & 0.581 & 0.691 & 0.648 & 0.733 \\
Qwen3-8B
& 61.60 & 2.78 & 26.24 & \textbf{0.593} & 0.693 & 0.670 & 0.757 \\
Qwen3-14B
& 61.78 & 1.15 & 29.90 & 0.584 & 0.681 & \textbf{0.686} & \textbf{0.780} \\
Qwen3-30B-A3B
& 53.29 & 5.92 & 28.22 & 0.587 & \textbf{0.695} & 0.646 & 0.729 \\
Gemma4-E4B
& 56.43 & 13.10 & 44.86 & 0.581 & 0.686 & 0.670 & 0.763 \\
Gemma4-12B
& 60.43 & 9.72 & 35.97 & 0.582 & 0.685 & 0.620 & 0.714 \\
Gemma4-26B-A4B
& \textbf{62.15} & 8.27 & 30.04 & 0.578 & 0.689 & 0.624 & 0.722 \\
\bottomrule
\end{tabular}
\end{table}

\begin{table}[h]
\centering
\vspace{-0.2in}
\caption{\textbf{Source-task and cross-benchmark performance after 2Wiki supervision.}
All systems use Qwen3-8B. Trained variants use only 2Wiki examples for gradient updates.}
\label{tab:2wiki-transfer}
\small
\setlength{\tabcolsep}{8pt}
\begin{tabular}{lcccccccc}
\toprule
& \multicolumn{2}{c}{\shortstack{\textbf{2Wiki}\\(Source)}}
& \multicolumn{2}{c}{\shortstack{\textbf{MuSiQue}\\(Transfer)}}
& \multicolumn{2}{c}{\shortstack{\textbf{HotpotQA}\\(Transfer)}}
& \multicolumn{2}{c}{\shortstack{\textbf{QuALITY}\\(Transfer)}}\\
\cmidrule(lr){2-3}\cmidrule(lr){4-5}
\cmidrule(lr){6-7}\cmidrule(lr){8-9}
\textbf{Method}
& \textbf{EM}$\uparrow$ & \textbf{F1}$\uparrow$
& \textbf{EM}$\uparrow$ & \textbf{F1}$\uparrow$
& \textbf{EM}$\uparrow$ & \textbf{F1}$\uparrow$
& \textbf{Acc.}$\uparrow$ & \textbf{H.Acc.}$\uparrow$ \\
\midrule
Raw Model
& 0.516 & 0.605 & 0.342 & 0.450
& 0.503 & 0.677 & 0.623 & 0.561 \\
Answer-only
& 0.698 & 0.775 & 0.482 & 0.579
& 0.630 & 0.763 & 0.537 & 0.465 \\
\midrule
Original Trace
& 0.599 & 0.676 & 0.387 & 0.486
& 0.589 & 0.736 & 0.620 & 0.567 \\
STaR
& 0.675 & 0.731 & 0.448 & 0.548
& 0.597 & 0.740 & 0.623 & 0.535 \\
NLL
& 0.677 & 0.763 & 0.463 & 0.561
& 0.616 & 0.762 & 0.607 & 0.522 \\
RSR
& 0.629 & 0.718 & 0.449 & 0.567
& 0.634 & 0.781 & 0.613 & 0.554 \\
LARK
& 0.684 & 0.769 & 0.469 & 0.561
& 0.625 & 0.768 & 0.577 & 0.490 \\
MBT-R
& 0.707 & 0.755 & 0.512 & 0.618
& 0.630 & \textbf{0.782} & 0.623 & 0.535 \\
\midrule
\textbf{\ReDiST (Ours)}
& \textbf{0.727} & \textbf{0.791}
& \textbf{0.536} & \textbf{0.633}
& \textbf{0.638} & \textbf{0.782}
& \textbf{0.660} & \textbf{0.580} \\
\bottomrule
\end{tabular}
\vspace{3pt}
\end{table}

\paragraph{Bootstrap intervals.}
Table~\ref{tab:qwen8b-both-directions} reports 95\% percentile bootstrap intervals for the \ReDiST Qwen3-8B point estimates, obtained by resampling the fixed 1,000-question evaluation slice with replacement 10,000 times. On a 1,000-question slice the intervals are roughly six points wide, which sets the resolution of the comparisons. On the source tasks, \ReDiST has the highest point estimate in both directions, but its interval contains the point estimates of the strongest baselines (MBT-R, RSR, and Answer-only under MuSiQue supervision; MBT-R under 2Wiki supervision), so the ordering among the top source-task recipes is not resolved at this sample size. The MuSiQue-to-2Wiki transfer margin is different in kind: no baseline point estimate falls inside the \ReDiST interval, and the 5.3-point gap over MBT-R exceeds the interval half-width. In the reverse direction, \ReDiST exceeds every baseline except MBT-R by more than the interval half-width, while the 2.4-point margin over MBT-R lies within it.

\begin{table}[h]
\centering
\caption{\textbf{Qwen3-8B source-task and transfer EM under MuSiQue and 2Wiki supervision.} Each entry reports EM on a fixed 1{,}000-question validation slice with its 95\% bootstrap interval.}
\label{tab:qwen8b-both-directions}
\small
\setlength{\tabcolsep}{6pt}
\resizebox{0.85\linewidth}{!}{%
\begin{tabular}{lcccc}
\toprule
& \multicolumn{2}{c}{MuSiQue-trained} & \multicolumn{2}{c}{2Wiki-trained} \\
\cmidrule(lr){2-3}\cmidrule(lr){4-5}
Supervision & MuSiQue (Source) & 2Wiki (Transfer) & 2Wiki (Source) & MuSiQue (Transfer) \\
\midrule
Raw model & 0.342 {\tiny[0.313, 0.370]} & 0.516 {\tiny[0.485, 0.547]} & 0.516 {\tiny[0.485, 0.547]} & 0.342 {\tiny[0.313, 0.370]} \\
Answer-only & 0.562 {\tiny[0.531, 0.592]} & 0.479  {\tiny[0.448, 0.510]} & 0.698  {\tiny[0.669, 0.726]} & 0.482  {\tiny[0.451, 0.513]} \\
Original trace & 0.426  {\tiny[0.396, 0.457]} & 0.521  {\tiny[0.490, 0.551]} & 0.599  {\tiny[0.568, 0.629]} & 0.387  {\tiny[0.357, 0.417]} \\
STaR & 0.510  {\tiny[0.479, 0.542]} & 0.613  {\tiny[0.583, 0.643]} & 0.675  {\tiny[0.646, 0.704]} & 0.448  {\tiny[0.417, 0.478]} \\
NLL & 0.546  {\tiny[0.515, 0.577]} & 0.600  {\tiny[0.570, 0.630]} & 0.677  {\tiny[0.648, 0.705]} & 0.463  {\tiny[0.433, 0.493]} \\
RSR & 0.568 {\tiny[0.537, 0.599]} & 0.624  {\tiny[0.594, 0.654]} & 0.629  {\tiny[0.599, 0.659]} & 0.449  {\tiny[0.418, 0.479]} \\
LARK & 0.539  {\tiny[0.508, 0.569]} & 0.630  {\tiny[0.600, 0.660]} & 0.684  {\tiny[0.655, 0.712]} & 0.469  {\tiny[0.438, 0.500]} \\
MBT-R & 0.579   {\tiny[0.548, 0.609]} & 0.633  {\tiny[0.602, 0.662]} & 0.707  {\tiny[0.677, 0.736]} & 0.512  {\tiny[0.479, 0.542]} \\
\midrule
\textbf{\ReDiST (Ours)} & \textbf{0.584} {\tiny[0.553, 0.615]} & \textbf{0.686} {\tiny[0.658, 0.714]} & \textbf{0.727} {\tiny[0.700, 0.754]} & \textbf{0.536} {\tiny[0.505, 0.567]} \\
\bottomrule
\end{tabular}}
\end{table}

\paragraph{Answer-loss weight.}
Table~\ref{tab:alpha-sweep} sweeps $\alpha$ with the selected targets and exposure of full \ReDiST{} held fixed; $\alpha=0$ is the \textsc{No Answer Auxiliary} variant of Table~\ref{tab:redist-ablation}. Source-task EM is highest at $\alpha=0$ (0.592) and varies by less than one point up to $\alpha=0.15$, whereas both transfer benchmarks peak at $\alpha=0.15$ (0.686 on 2Wiki, 0.633 on HotpotQA). Beyond $\alpha=0.15$, all three benchmarks decline, with 2Wiki EM falling monotonically to 0.579 at $\alpha=0.90$. Considering the trade-off between different metrics, we use $\alpha=0.15$ throughout.

\begin{table}[h]
\centering
\caption{\textbf{Answer-loss weight sweep.}}
\label{tab:alpha-sweep}
\small
\setlength{\tabcolsep}{10pt}
\begin{tabular}{cccc}
\toprule
$\alpha$ & \textbf{MuSiQue EM}$\uparrow$
& \textbf{2Wiki EM}$\uparrow$
& \textbf{HotpotQA EM}$\uparrow$ \\
\midrule
0.00 & \textbf{0.592} & 0.652 & 0.630 \\
0.15 & 0.584 & \textbf{0.686} & \textbf{0.633} \\
0.30 & 0.572 & 0.653 & 0.602 \\
0.60 & 0.575 & 0.616 & 0.587 \\
0.75 & 0.558 & 0.610 & 0.590 \\
0.90 & 0.535 & 0.579 & 0.567 \\
\bottomrule
\end{tabular}
\end{table}

\paragraph{Cross-student comparison.}
Table~\ref{tab:cross-setting-utility-2wiki} repeats the four-student comparison of Table~\ref{tab:cross-setting-utility} with 2Wiki supervision. \ReDiST achieves the highest reported 2Wiki EM as the source task in all four comparisons and the highest MuSiQue EM as the transfer task, with NLL matching its transfer EM for Gemma4-E2B (0.458). Table~\ref{tab:musique-transfer-qwen3-4b} reports all four benchmarks for Qwen3-4B under MuSiQue supervision. \ReDiST obtains the highest value in every column except the two 2Wiki columns, where NLL is highest. These results further support the applicability of the construction and distillation pipeline across the evaluated model families, scales, and source tasks, while also indicating that its benefits over competing strategies vary with the target student.

\begin{table}[h]
\centering
\caption{\textbf{Source-task and transfer performance across target students.} All trained variants receive 2Wiki supervision. Columns report 2Wiki source-task (S) and MuSiQue transfer (T) EM for each student.}
\label{tab:cross-setting-utility-2wiki}
\small
\setlength{\tabcolsep}{8pt}
\begin{tabular}{lcccccccc}
\toprule
& \multicolumn{2}{c}{Qwen3-4B} & \multicolumn{2}{c}{Qwen3-8B} & \multicolumn{2}{c}{Gemma4-E2B} & \multicolumn{2}{c}{Gemma4-E4B} \\
\cmidrule(lr){2-3}\cmidrule(lr){4-5}\cmidrule(lr){6-7}\cmidrule(lr){8-9}
Supervision & S & T & S & T & S & T & S & T \\
\midrule
Raw model & 0.225 & 0.293 & 0.516 & 0.342 & 0.518 & 0.225 & 0.643 & 0.425 \\
Answer-only & 0.678 & 0.425 & 0.698 & 0.482 & 0.628 & 0.401 & 0.687 & 0.458 \\
Original trace & 0.576 & 0.400 & 0.599 & 0.387 & 0.554 & 0.264 & 0.661 & 0.468 \\
STaR & 0.644 & 0.446 & 0.675 & 0.448 & 0.644 & 0.430 & 0.759 & 0.546 \\
NLL & 0.659 & 0.423 & 0.677 & 0.463 & 0.669 & \textbf{0.458} & 0.692 & 0.557 \\
RSR & 0.635 & 0.408 & 0.629 & 0.449 & 0.672 & 0.438 & 0.687 & 0.553 \\
LARK & 0.686 & 0.401 & 0.684 & 0.469 & 0.668 & 0.441 & 0.701 & 0.553 \\
MBT-R & 0.666 & 0.366 & 0.707 & 0.512 & 0.653 & 0.435 & 0.691 & 0.555 \\
\textbf{\ReDiST (Ours)} & \textbf{0.709} & \textbf{0.479} & \textbf{0.727} & \textbf{0.536} & \textbf{0.704} & \textbf{0.458} & \textbf{0.762} & \textbf{0.574} \\
\bottomrule
\end{tabular}
\vspace{-0.1in}
\end{table}

\begin{table}[h]
\centering
\caption{\textbf{Source-task and cross-benchmark performance after MuSiQue supervision.} All systems use Qwen3-4B. Trained variants use only MuSiQue examples for gradient updates. Bold: best value in each column.}
\label{tab:musique-transfer-qwen3-4b}
\small
\setlength{\tabcolsep}{8pt}
\begin{tabular}{lcccccccc}
\toprule
& \multicolumn{2}{c}{\shortstack{\textbf{MuSiQue}\\(Source)}}
& \multicolumn{2}{c}{\shortstack{\textbf{2Wiki}\\(Transfer)}}
& \multicolumn{2}{c}{\shortstack{\textbf{HotpotQA}\\(Transfer)}}
& \multicolumn{2}{c}{\shortstack{\textbf{QuALITY}\\(Transfer)}}\\
\cmidrule(lr){2-3}\cmidrule(lr){4-5}\cmidrule(lr){6-7}\cmidrule(lr){8-9}
\textbf{Method} & \textbf{EM}$\uparrow$ & \textbf{F1}$\uparrow$ & \textbf{EM}$\uparrow$ & \textbf{F1}$\uparrow$ & \textbf{EM}$\uparrow$ & \textbf{F1}$\uparrow$ & \textbf{Acc.}$\uparrow$ & \textbf{H.Acc.}$\uparrow$ \\
\midrule
Raw Model & 0.293 & 0.421 & 0.225 & 0.413 & 0.407 & 0.600 & 0.560 & 0.510 \\
Answer-only & 0.558 & 0.676 & 0.440 & 0.521 & 0.547 & 0.692 & 0.470 & 0.395 \\
\midrule
Original Trace & 0.405 & 0.500 & 0.445 & 0.534 & 0.586 & 0.730 & 0.587 & 0.535 \\
STaR & 0.519 & 0.618 & 0.480 & 0.558 & 0.606 & 0.746 & 0.527 & 0.497 \\
NLL & 0.519 & 0.621 & \textbf{0.635} & \textbf{0.723} & 0.606 & 0.755 & 0.597 & 0.522 \\
RSR & 0.533 & 0.642 & 0.587 & 0.679 & 0.604 & 0.755 & 0.570 & 0.490 \\
LARK & 0.507 & 0.610 & 0.626 & 0.709 & 0.619 & 0.766 & 0.557 & 0.516 \\
MBT-R & 0.556 & 0.657 & 0.625 & 0.710 & 0.615 & 0.760 & 0.557 & 0.510 \\
\midrule
\textbf{\ReDiST (Ours)} & \textbf{0.580} & \textbf{0.706} & 0.630 & 0.717 & \textbf{0.620} & \textbf{0.777} & \textbf{0.613} & \textbf{0.548} \\
\bottomrule
\end{tabular}%
\vspace{-0.05in}
\end{table}

\clearpage
\section{Prompts}

In this section, we provide the prompts used in our experiments. The student prompt below is the raw-text form used for Qwen students, which ends inside an open \texttt{<think>} block. Gemma students receive the same instruction (except for the abstention option), question, and evidence as the user turn of their chat template with thinking disabled; the tagged output format is unchanged. The reader prompt is wrapped in the same way as the student prompt of the corresponding model family, and the teacher prompts are sent as a single user turn of the rewrite model's chat template.

\begin{tcolorbox}[
  title={\ReDiST{} Student Prompt (Draft Generation, Training, and Evaluation)},
  colback=white,
  colframe=black,
  boxrule=0.5pt,
  breakable
]
\begin{Verbatim}[fontsize=\small, breaklines=true]
You are a precise multi-hop QA assistant.
Read the EVIDENCE and answer the QUESTION.
You MUST reason in multiple concise steps before final answer.
Output format must be EXACTLY:
<think>
1) step one
2) step two
...
</think>
<answer>short final answer phrase</answer>
No text outside these tags.
If the EVIDENCE is insufficient to determine the answer, give UNANSWERABLE as the final answer.

QUESTION:
<question>

EVIDENCE:
Title: <title_1>
<paragraph_1>

Title: <title_2>
<paragraph_2>
...

OUTPUT:
<think>
\end{Verbatim}
\end{tcolorbox}

\begin{tcolorbox}[
  title={\ReDiST{} Teacher Rewrite Prompt (Correct Draft)},
  colback=white,
  colframe=black,
  boxrule=0.5pt,
  breakable
]
\begin{Verbatim}[fontsize=\small, breaklines=true]
The student's draft reached the correct answer. Improve its clarity and evidential support without changing the supported conclusion.
Rewrite the reasoning into a complete evidence-grounded explanation. You may reorder, replace, add, or remove steps and change the wording; you do not need to minimize edits or preserve the original length.
Resolve the intermediate entities and explicitly connect the relevant facts from the evidence to the final answer-bearing fact. Keep useful reasoning, remove unsupported claims and irrelevant details. Use only the evidence, not outside knowledge. Prefer a concise complete chain, usually 3-7 numbered steps; do not pad easy questions.
State the final relationship as a natural factual statement. Do not write 'the answer is', repeat an answer hint, mention a supplied/gold answer, the draft, or the rewriting instructions. Output ONLY the reasoning, with no think/answer tags or separate answer declaration.

QUESTION:
<question>

EVIDENCE DOCUMENTS:
<evidence>

DRAFT REASONING:
<student_draft_reasoning>

REWRITTEN REASONING:
\end{Verbatim}
\end{tcolorbox}

\begin{tcolorbox}[
  title={\ReDiST{} Teacher Repair Prompt (Incorrect Draft)},
  colback=white,
  colframe=black,
  boxrule=0.5pt,
  breakable
]
\begin{Verbatim}[fontsize=\small, breaklines=true]
The student's draft reached a wrong answer. Correct the faulty reasoning and derive the supplied target from the evidence. The target is a checking hint, not evidence; do not invent facts to justify it.
Rewrite the reasoning into a complete evidence-grounded explanation. You may reorder, replace, add, or remove steps and change the wording; you do not need to minimize edits or preserve the original length.
Resolve the intermediate entities and explicitly connect the relevant facts from the evidence to the final answer-bearing fact. Keep useful reasoning, remove unsupported claims and irrelevant details. Use only the evidence, not outside knowledge. Prefer a concise complete chain, usually 3-7 numbered steps; do not pad easy questions.
State the final relationship as a natural factual statement. Do not write 'the answer is', repeat an answer hint, mention a supplied/gold answer, the draft, or the rewriting instructions. Output ONLY the reasoning, with no think/answer tags or separate answer declaration.

QUESTION:
<question>

EVIDENCE DOCUMENTS:
<evidence>

DRAFT REASONING:
<student_draft_reasoning>

CORRECT FINAL ANSWER:
<gold_answer>

REWRITTEN REASONING:
\end{Verbatim}
\end{tcolorbox}

\begin{tcolorbox}[
  title={\ReDiST{} Reader Prompt (Candidate Selection and Verification)},
  colback=white,
  colframe=black,
  boxrule=0.5pt,
  breakable
]
\begin{Verbatim}[fontsize=\small, breaklines=true]
<student prompt above, ending with "OUTPUT:\n<think>\n">
<candidate reasoning (original draft or teacher rewrite)>
</think>
<answer>
\end{Verbatim}
\end{tcolorbox}

\end{document}